\documentclass{article}

   \usepackage[nonatbib, preprint]{neurips_2020}

\usepackage{graphicx}
\usepackage{amsmath}
\usepackage{wrapfig}

\usepackage[utf8]{inputenc} 
\usepackage[T1]{fontenc}    
\usepackage{hyperref}       
\usepackage{url}            
\usepackage{booktabs}       
\usepackage{amsfonts}       
\usepackage{nicefrac}       
\usepackage{microtype}      
\usepackage{placeins}

\title{Formatting Instructions For NeurIPS 2020}

\author{%
 Qendrim Bytyqi\\
 \textit{Hochschule für Technik Stuttgart} \\
 \textit{University of Applied Sciences}\\
 Stuttgart, Germany \\
 qendrim.bytyqi@hft-stuttgart.de\\
  \And
  Nicola Wolpert\\
  \textit{Hochschule für Technik Stuttgart} \\
  \textit{University of Applied Sciences}\\
  Stuttgart, Germany \\
  nicola.wolpert@hft-stuttgart.de\\
  \And
  Elmar Schömer\\
  \textit{Johannes Gutenberg University Mainz} \\
  \textit{Institute of Computer Science}\\
  Mainz, Germany \\
  schoemer@informatik.uni-mainz.de
}

\begin{document}

\title{Local-Area-Learning Network: Meaningful Local Areas for Efficient Point Cloud Analysis}

\maketitle

\begin{abstract}
	
	Research in point cloud analysis with deep neural networks has made rapid progress in recent years. The pioneering work PointNet offered a direct analysis of point clouds. However, due to its architecture PointNet is not able to capture local structures. To overcome this drawback, the same authors have developed PointNet++ by applying PointNet to local areas. The local areas are defined by center points and their neighbors. In PointNet++ and its further developments the center points are determined with a Farthest Point Sampling (FPS) algorithm. This has the disadvantage that the center points in general do not have meaningful local areas. In this paper, we introduce the neural Local-Area-Learning Network (LocAL-Net) which places emphasis on the selection and characterization of the local areas. Our approach learns critical points that we use as center points. In order to strengthen the recognition of local structures, the points are given additional metric properties depending on the local areas. Finally, we derive and combine two global feature vectors, one from the whole point cloud and one from all local areas. Experiments on the datasets ModelNet10/40 and ShapeNet show that LocAL-Net is competitive for part segmentation. For classification LocAL-Net outperforms the state-of-the-arts.
	
	
\end{abstract}

\section{INTRODUCTION}

The analysis of 3D point clouds has gained a growing importance due to its application in autonomous driving and scene understanding. However, the use of Deep Learning methods based on point clouds is very challenging because point clouds are sparse and unordered. Since the neural network must be independent of the input order of the points, common architectures like CNNs that require a structured input cannot be applied directly to point clouds. Some approaches transform the point cloud into a structured data format like a voxel grid \cite{voxandmul, mn} so that 3D-CNNs \cite{3dcnn} can be used. This procedure unnecessarily increases the complexity and the memory usage and usually distorts the shape of the objects.

PointNet solves the problems by applying a shared multi layer perceptron (MLP) to each point individually and by aggregating the resulting information through a symmetric function across all points. However, PointNet ignores local structures, which is why further developments look at neighborhoods of points and do not process them individually. DGCNN \cite{dgcnn} constructs a local graph for each point and its neighbors and learns edge features that describe the relationship between them. PointNet++ \cite{pn++} defines local areas by center points and their neighbors. It learns on different scales of these local areas and produces via an abstraction layer a point set with fewer elements than the input point set. This procedure is applied multiple times to get a hierarchical learning procedure. Point2Sequence \cite{p2s} uses a RNN-based model with an attention mechanism to capture the correlation of different scales in local regions. By now, all these approaches consider local structures that are not necessarily geometrically meaningful. For example, PointNet++ and Point2Sequence determine the local areas by using a deterministic Farthest Point Sampling (FPS) algorithm to find center points and by computing $k$-nearest-neighbors ($k$-NN) around them. Thus the local areas are evenly distributed and the neural network has to deal with partly insignificant or redundant geometric information which can make the capturing of important features difficult.

\begin{figure*}
	\centering
	\includegraphics[width=1.0\linewidth]{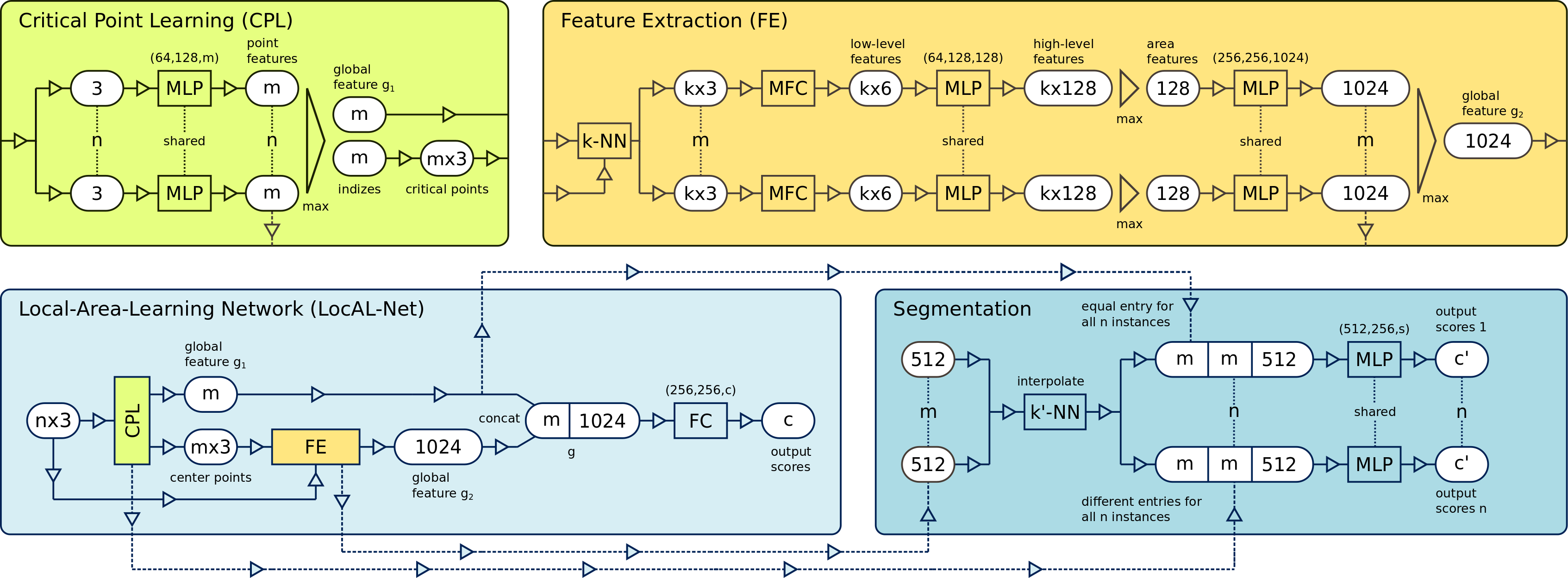}
	\caption{\textbf{LocAL-Net architecture.} The network first determines in CPL center points from the input point cloud and generates a global feature vector $g_1$. In FE local areas are generated and used to generate area features and a second global feature vector $g_2$. The concatenation $g$ of $g_1$ and $g_2$ in LocAL-Net can be used to generate an output score for classification. The segmentation network is an extension of the classification network with slight modifications.}
	\label{fig:netz}
\end{figure*}

\begin{samepage}
We adress this problem by proposing the neural network LocAL-Net which contains the following three main ideas, consider also Figure \ref{fig:netz}:
\begin{itemize}
	\item[(1)] The Critical-Point-Learning (CPL) sub-network learns to find $m$ critical points with a characteristic neighborhood and uses them as center points. It generates a first global feature vector $g_1$ by considering the whole point cloud.
	
	\item[(2)] The Feature-Extraction (FE) sub-network enriches the $m$ center points in their respective local neighborhoods with low-level features, which enables a better grasping of the underlying geometry in the local areas. It generates a second global feature vector $g_2$ from all local areas.
	
	\item[(3)] The combination of the two global feature vectors $g_1$ and $g_2$ from different scale considerations enables LocAL-Net to recognize the correlation of the local areas.
\end{itemize}
Since the learned points from the Critical-Point-Learning (CPL) part (1) are used as center points, LocAL-Net can capture important features of the object without being hindered by insignificant information. In the Feature-Extraction (FE) part (2), we first determine for each center point a neighbourhood by computing $k$-NN. Then, in Metric-Feature-Computation (MFC), each of the $k$ points is enriched by permutation-invariant low-level features that depend on its respective neighborhood. We use metric properties which enable LocAL-Net to capture well the underlying geometry of the local areas. The resulting low-level feature vector of each point is converted into a high-level feature vector. For each local area the $k$ informations are then aggregated into an area feature vector by a symmetric function. We apply this procedure again to the area feature vectors and obtain the second global feature vector $g_2$. In the overall LocAL-Net~(3) we finally concatenate the two global feature vectors $g_1$ and $g_2$, which were obtained from different scales of the input point cloud.
Experimental results show that LocAL-Net learns to find center points that are characteristic for the given point cloud. Experiments on the datasets ModelNet10/40 and ShapeNet show that LocAL-Net is competitive for part segmentation. For classification LocAL-Net outperforms the state-of-the-arts.

\section{Related Work}

Deep learning approaches for 3D object analysis can be divided into three main classes.

\textbf{Volumetric methods} approximate the input data with voxels so that extensions of conventional CNNs to 3D data can be used \cite{voxandmul, mn, 3dcnn, learnvox}. The main disadvantage of these approaches is the time effort for the conversion and the representation quality. Some approaches \cite{voxfpnn, kdtree, octnet} propose special methods to solve the sparsity problem of a volumetric representation. For example, OctNet~\cite{octnet} divides the space hierarchically, depending on the input, by using a set of unbalanced octrees where each leaf node stores a bundled feature representation. This makes it possible to concentrate memory allocation and calculation on the relevant dense regions. However, to overcome the problem with the representation quality, a high resolution of the voxel grid is required. This leads to large memory consumption and a high running time for large point clouds.

\textbf{Multiview-based methods} follow the idea of rendering 3D CAD data into 2D images from different views to take advantage of the power of traditional CNNs \cite{mvcnn, rotnet}. Thanks to the rapid development of 2D CNNs for images in recent years, such methods have achieved dominant results in classification and retrieval tasks for 3D CAD data. However, it is a nontrivial problem to extend the results from surface objects to discrete point clouds.
\end{samepage}

\textbf{Deep Learning directly on point clouds} was first discussed by PointNet \cite{pn}. The authors describe the irregular format and permutation invariance problem of point clouds and present a neural network that directly acts on point clouds. The main idea to solve the permutation invariance problem is to transform a point cloud $\{p_1,\dots,p_n\}\subset\mathbb{R}^d$ by applying a shared MLP $h$ to all points and aggregating the information with a symmetric function $g$:
\[f(p_1,\dots,p_n)=g(h(p_1),\dots,h(p_n)).\]
The same authors have developed PointNet++ \cite{pn++} by applying PointNet to local areas. The results have shown that the properties of local areas improve deep learning on 3D point clouds. The local areas are defined by center points and their neighbors. In PointNet++ and its further developments \cite{pweb, p2s, rscnn} the center points are determined with a Farthest Point Sampling (FPS) algorithm. This has the disadvantage that the center points are not necessarily important points with a meaningful neighborhood. DGCNN \cite{dgcnn} constructs a local graph for each point of the point cloud by connecting the point to its neighbors. It learns edge features to capture local details. This basic procedure is called EdgeConv. By stacking EdgeConv layers, the network can learn global shape properties. However, there is the same problem as with PointNet++: Treating all points and therefore all areas equally makes the capturing of important features difficult.

\section{Our Approach}

In this section we want to give an overview of the setup and the architecture of our neural network LocAL-Net. After that, we will discuss the components in detail.

\subsection{Setup}

The input point cloud $P\subset \mathbb{R}^3$ is an unordered point set that is defined as \[P:=\{p_i=(x_i,y_i,z_i)~|~i=1,\dots,n\},\] where $x_i,y_i,z_i$ are the coordinates of $p_i$. It is conceivable that additional features may be given for each point such as color, normal or manually calculated characteristics. Therefore, we generalize this definition as follows:
\[
P^{(l)}:=\left\{
\begin{array}{ll}
P & \mbox{, } l = 0 \\
\{(x_i,y_i,z_i, \phi_{i_1},\dots,\phi_{i_l})~|~i=1,\dots,n\} &  \mbox{, else} \\
\end{array}
\right.
\]
for $l\in\mathbb{N}$ and $P^{(l)}\subset \mathbb{R}^{3+l}$. 

In this paper we are dealing with two types of tasks: classification and part segmentation. For the classification task we want to classify the whole point cloud $P$. Let $c_1,\dots,c_m$ be the different classes. The neural network $F$ should convert the input point cloud $P$ to a probability distribution
\[F(P)=\left(Pr(c_1|P),\dots, Pr(c_m|P)\right).\]

In the part segmentation task, we need to classify each point $p_i\in P$ into sub-classes $\tilde{c}_1,\dots,\tilde{c}_s$ of the objects. The output of $F(P)$ contains for each point $p_i\in P$ a probability distribution
\begin{equation}\label{seg}
F(P)=\left( \begin{array}{rrrr} Pr_{p_1}(\tilde{c}_1|P) & \cdots & Pr_{p_1}(\tilde{c}_s|P) \\ \vdots \hspace{0.8cm} &  &  \vdots\hspace{0.8cm} \\Pr_{p_n}(\tilde{c}_1|P) & \cdots & Pr_{p_n}(\tilde{c}_s|P) \\\end{array}\right).
\end{equation}

\subsection{Architecture}

Figure \ref{fig:netz} shows the architecture of our model. The input of LocAL-Net is a point cloud $P=~\{p_1,\dots,p_n\}$. The CPL part extracts $m$ critical points, some of which can occur several times, and a first global feature vector $g_1$ from $P$. We use the critical points as center points. In FE we define a local area for each center point with its $k$-NN points in $P$.
\begin{wrapfigure}{R}{7.5cm}
	\vspace{-1.0cm}
	\includegraphics[width=7.5cm]{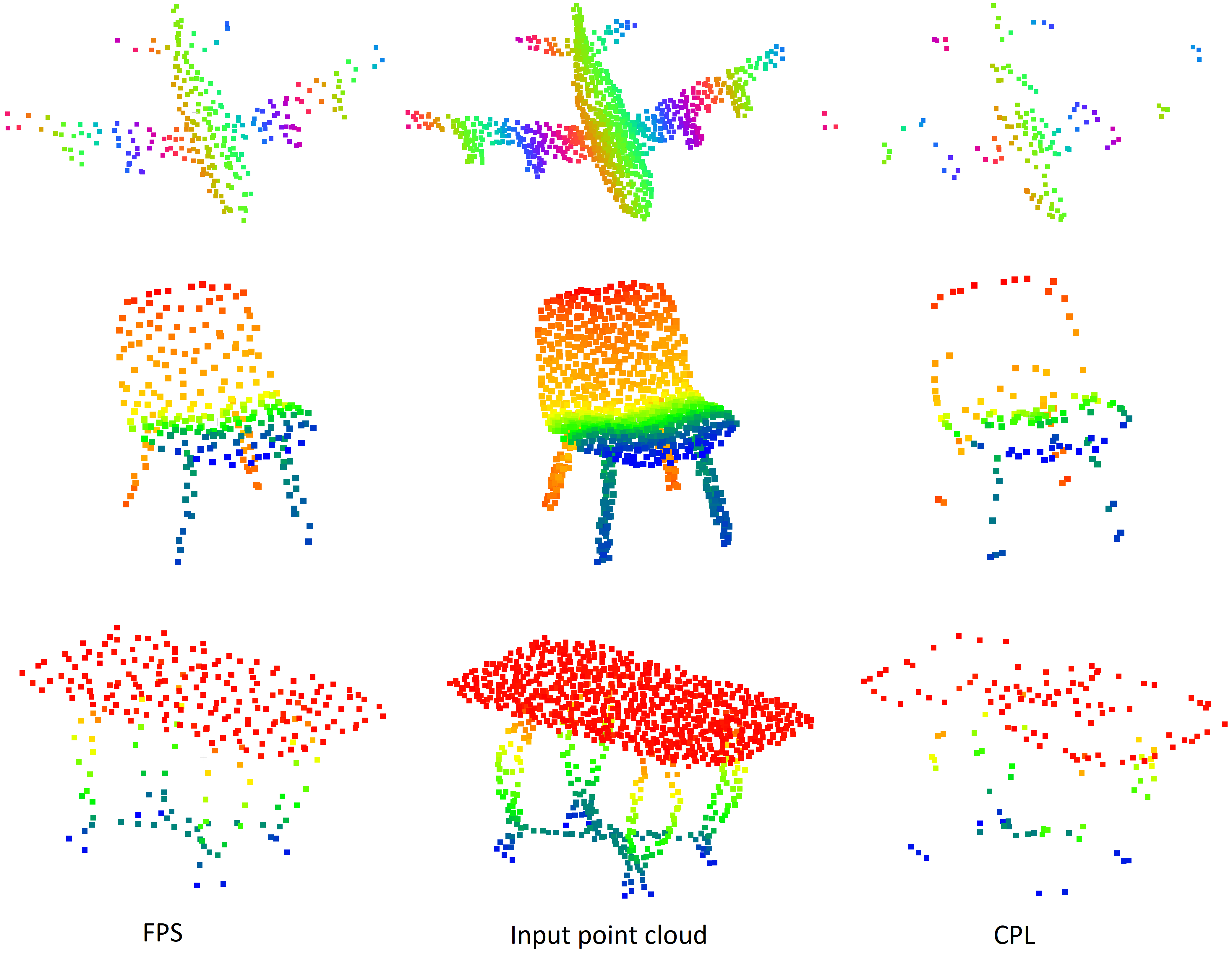}
	\caption{\textbf{Comparison of center points created with FPS and CPL.} While FPS creates a uniformly distributed subset (left column) of the input point cloud (middle column), CPL selects center points with a meaningful neighborhood (right column).}
	\label{fig:stuhl}
\end{wrapfigure}
Based on these local areas, FE generates in a local to global manner the second global feature vector $g_2$. Finally, in LocAL-Net $g_1$ and $g_2$ are concatenated to obtain a multi scale global feature vector $g$,which can be used for shape classification and, with slight modifications, for part segmentation.

\subsection{Critical Point Learning (CPL)}

For the given point cloud $P$, every point $p_i \in P$ is mapped with a shared MLP into an $m$-dimensional feature vector. The first $m$-dimensional global feature vector $g_1$ is created by applying max pooling over all these $n$ feature vectors. The authors of PointNet noted that by applying max pooling over the feature vectors of the points, the network allows important points to participate preferentially in the global feature vector $g_1$. For each entry of $g_1$ we determine the index of the point that is responsible for the entry. Every index defines a critical point. Consequently, there can be at most $m$ pairwise different critical points. We use these critical points $P'=\{p_1',\dots,p_m'\}\subset P$ as center points. An advantage of CPL over FPS for computing the center points is the fact that the required calculations are matrix operations that are highly efficient on the GPU. FPS is an iterative algorithm that does not allow parallel computation. Even more important we can see in Fig.~\ref{fig:stuhl} that LocAL-Net learns critical points with geometrically meaningful neighborhoods. The middle column of Figure~\ref{fig:stuhl} shows three input point clouds. On the left one can see the center points computed with FPS and on the right the center points learned by CPL. The point set obtained by FPS is evenly distributed and therefore every area, important or not, is considered. The point set learned by LocAL-Net, on the other hand, emphasizes characteristic areas. For the aircraft, points on the wing tips, turbines and the fuselage are selected. Points on areas like the flat surfaces of the wings are not chosen. This can also be observed with the chair and the table. Points in flat areas are rarely picked. CPL preferably selects points on the contour of the seat and the table top. Also extremity points on the legs of the chair and the table are learned. The conclusion is that
\begin{itemize}
	\item[i)] including the definition of the center points into the learning process and
	\item[ii)] restricting the number of center points to a small fraction of the number of input points 
\end{itemize}
gives LocAL-Net the ability to learn a few but meaningful center points. Our new CPL computation can also be used to replace the FPS in other neural networks that rely on center points.

\subsection{Feature Extraction (FE)}

\begin{wrapfigure}{r}{7.5cm}
	\vspace{-0.2cm}
	\label{fig:bereiche}
	\includegraphics[width=7.5cm]{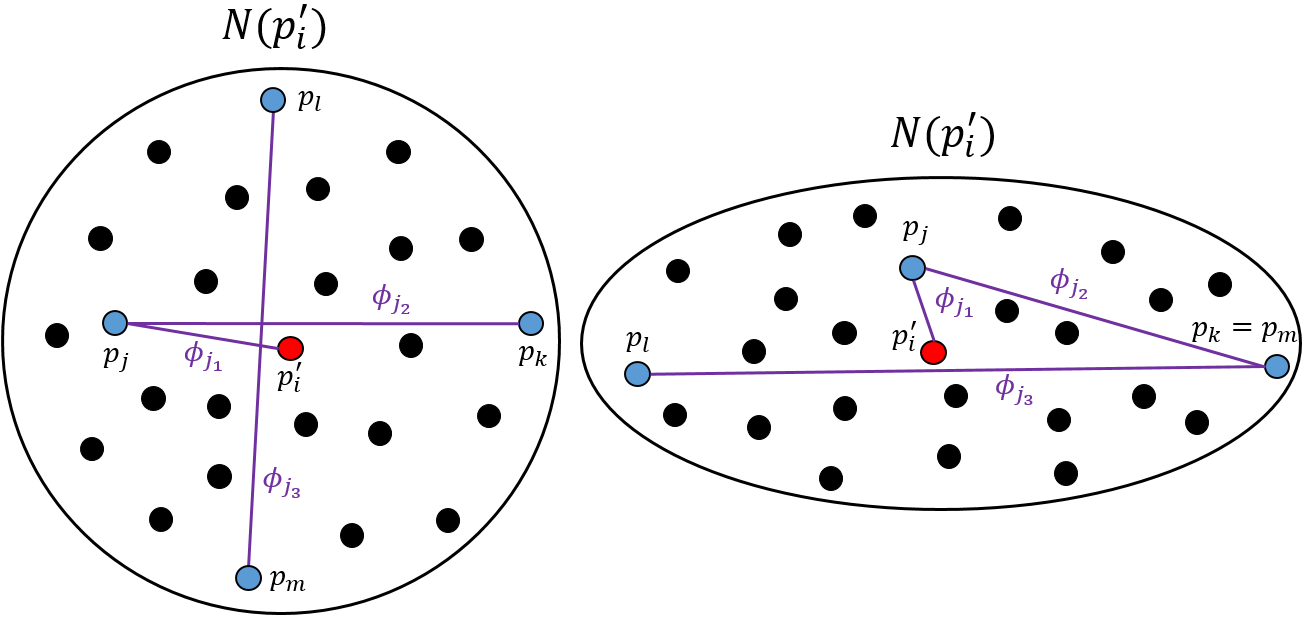}
	\caption{\textbf{Illustration of the metric features} for two different point sets in the 2D Euclidean space.}
	\vspace{-0.7cm}
\end{wrapfigure}

For each center point $p'_i\in P'$ we compute its local area $N(p_i')\subset P$ by determining its $k$-NN. Similar to PointNet++ the points $p_j$ in a local area $N(p_i')$ are translated into a relative coordinate system of its corresponding center point $p_i'$ by $p_j=p_j-p_i'$. In the Metric-Feature-Computation (MFC) part we calculate for each point $p_j=(x_j,y_j,z_j)\in N(p_i')$ the following metric properties:
\begin{align*}
\phi_{j_1}:=\|p_j\|_2,\quad\quad\quad
\phi_{j_2}:=\underset{p_k\in N(p_i')}{\max}\|p_j-p_k\|_2,\quad\quad\quad \phi_{j_3}:=\underset{p_l,p_m\in N(p_i')}{\max}\|p_l-p_m\|_2.
\end{align*}
Fig. \ref{fig:bereiche} illustrates the metric features for 2D points in the Euclidean plane. The values $\phi_{j_1}$ and $\phi_{j_2}$ help LocAL-Net to find out where the location of $p_j$ in $N(p_i')$ is. $\phi_{j_3}$ has a constant value for all points in $N(p_i')$. It could be used by LocAL-Net as an information about the shape of $N(p_i')$. Since we always consider $k$-NN for a constant $k$, a small value of $\phi_{j_3}$ indicates a circle like area on the surface wheres a large value implies a narrow and oblong area. The properties are added to $p_j$ which leads to a low-level feature vector \[\tilde{p}_j:=~(x_j,y_j,z_j,\phi_{j_1},\phi_{j_2},\phi_{j_3})\in P^{(3)}.\] 
The low-level feature vectors are transformed into high-level feature vectors by using a shared MLP. We then extract an area feature vector for each area $N(p_i')$ through a symmetric function. To aggregate the informations of the areas we apply this procedure again and obtain a second global feature vector $g_2$ for the entire input point cloud. 

\subsection{LocAL-Net}

Finally, in the overall LocAL-Net we concatenate the two global feature vectors $g_1$ and $g_2$ which leads to a multiscale global feature vector $g$. Both vectors, $g_1$ and $g_2$ are crucial. Only through $g_1$ the parameters in CPL are learned in the training procedure by the backpropagation algorithm. The parameters themselves determine the critical points. Without $g1$, the critical points would be randomly selected. $g_2$ contains features obtained from a local to global manner, which are important for analyzing point clouds as PointNet++ has shown. Without $g_2$, we would have a simple PointNet (Vanilla) architecture.

\section{Experiments}

In this section, we evaluate LocAL-Net on the shape classification and part segmentation tasks. In the shape classification task we also perform various tests on ModelNet40 with different hyperparameters and study the effect of the CPL and FE part. For each test scenario we change one parameter and carry out five runs and take the best result. All experiments are performed on an NVIDIA GeForce RTX 2080 Ti GPU and implemented using PyTorch. Our code and models will soon be publicly available.

\subsection{3D Shape Classification}

For the 3D Shape Classification task we evaluate LocAL-Net on the ModelNet10 and the ModelNet40 benchmarks \cite{mn}. The ModelNet10 dataset contains 4,899 meshed CAD models from 10 different categories. Similar to \cite{p2s} we split the dataset into 3,991 samples for training and 908 samples for testing. The ModelNet40 dataset contains 12,311 meshed CAD models from 40 different categories. Similar to \cite{pn, pn++} we split the dataset into 9,843 samples for training and 2,468 samples for testing. We uniformly sample 1,024 points on the mesh of each CAD model and normalize them into a unit sphere. During training we augment these input point clouds with anisotropic scaling in the range $[0.66,1.4]$ and translation in the range $[-0.2,0.2]$ as in \cite{kdtree, rscnn}. Furthermore we add a Gaussian noise with $0$ mean and $0.01$ standard deviation as in \cite{pn, pn++}. All layers of our model, except the last one, are followed by ReLU and batch normalization with $0.1$ momentum. For the two layers before score prediction we use dropout with $0.5$ ratio. The Adam optimizer \cite{adam} with batch size $16$ and initial learning rate $0.001$ is used to optimize the cross entropy loss function during training. The learning rate decays every $23$ epochs with a rate of $0.7$. Similar to \cite{rscnn} we do a voting with 10 anisotropic scaled point clouds in the test.

The results are shown in Table~\ref{table:modelnet}. LocAL-Net achieves the best results of all point cloud based methods on both datasets. On the ModelNet10 dataset our model reduces the error rate of SO-Net by $14.0\%$ although we only use 1,024 points with $xyz$ coordinates. Compared to the networks that equally only uses 1,024 points with only $xyz$ coordinates, LocAL-Net reduces the error of the current best result from Point2Seq by $21.3\%$. On the ModelNet40 dataset LocAL-Net archives a highest accuracy of $93.7\%$ with voting and $93.3\%$ without voting which reduces the error by $1.6\%$ and $5.6\%$ of the current best results archived by RS-CNN with and without voting. To evaluate the stability of LocAL-Net we trained 20 runs on ModelNet40. Then we computed the mean test accuracy with voting and got a value of $93.3\%$. The training of 200 epochs on ModelNet40 took about 4 hours.

\begin{table}[]
	\caption{Results of the shape classification on ModelNet10 and ModelNet40.}
	\label{table:modelnet}
	\centering
		\begin{tabular}{l|c|c|c|c}
			\hline
			Method        & Input    & \#points & \begin{tabular}[c]{@{}c@{}}ModelNet10 \\ Acc (\%)\end{tabular} & \begin{tabular}[c]{@{}c@{}}ModelNet40\\ Acc (\%)\end{tabular} \\ \hline
			Pointwise-CNN \cite{pointwise}& xyz      & $1024$   &                                                            & $86.1$                                                    \\
			PointNet \cite{pn}     & xyz      & $1024$   & -                                                          & $89.2$                                                    \\
			Kd-Net \cite{kdtree}      & xyz      & $1024$   & $93.3$                                                     & $90.6$                                                    \\
			PointNet++\cite{pn++}    & xyz      & $1024$   & -                                                          & $90.7$                                                    \\
			KC-Net  \cite{kcnet}   & xyz      & $1024$   &    
			$94.4$                                                     & $91.0$                                                    \\
			Pointweb  \cite{pweb}   & xyz      & $1024$   &    
			-                                                          & $92.3$                                                    \\
			PointCNN \cite{DBLP:journals/corr/abs-1801-07791}     & xyz      & $1024$   & -                                                          & $92.5$                                                    \\
			Point2Seq  \cite{p2s}   & xyz      & $1024$   & $95.3$                                                     & $92.6$                                                    \\
			3DCapsule   \cite{3dcapsule}  & xyz      & $1024$   & $94.7$                                                     & $92.7$                                                    \\
			DGCNN   \cite{dgcnn}      & xyz      & $1024$   & -                                                          & $92.9$                                                    \\
			RS-CNN   \cite{rscnn}     & xyz      & $1024$   & -                                                          & $92.9$                                                    \\ 
			RS-CNN (voting)  \cite{rscnn}     & xyz      & $1024$   & -                                                          & $93.6$                                                    \\ \hline
			Ours                     & xyz       & $1024$   & \textbf{95.8}                                                    &   \textbf{93.3}                                        \\ \hline
			Ours (voting)         & xyz      & $1024$   & \textbf{96.3}                                            & \textbf{93.7}                                           \\ \hline
			PointNet++\cite{pn++}    & xyz, nor & $5000$   & -                                                          & $91.9$                                                    \\ 
			SpiderCNN  \cite{spidercnn}   & xyz, nor & $5000$   & -                                                          & $92.4$                                                    \\ 
			SO-Net   \cite{sonet}     & xyz, nor & $5000$   & $95.7$                                                     & $93.4$                                                   
		\end{tabular}
\end{table}

\subsection{CPL tests}

\textbf{Determine center points.} In the first test we replace the complete CPL sub-network by the classical FPS to determine the center points. The results are shown in Table \ref{table:cplfps}. The acceptance rate drops from $93.7\%$ to $93.2\%$. This comparison reinforces our hypothesis that significant local areas support learning and generalization in a neural network. In this first test also the global feature vector g1 (as part of CPL) is removed. In a second test we investigate the effect of g1. We use FPS to determine the center points but keep $g_1$ for classification. This increases the test accuracy to $93.5\%$. This second result suggests that LocAL-Net is able to consider global properties to effectively analyze the input point cloud.

\begin{table}[!h]
	\caption{The results of FPS and CPL on ModelNet40}
	\label{table:cplfps}
	\centering
	\begin{tabular}{l|c|c|c}
		Method   & FPS without $g_1$ & FPS with $g_1$ & CPL \\ \hline
		Acc (\%) & $93.2$ & $93.5$  & $93.7$  
	\end{tabular}
\end{table}

\textbf{Number of critical points.} In the next test shown in Table~\ref{table:center} we change the parameter $m$ from $192$ to $320$ with a step size of $32$. The comparison shows that the instance accuracy increases fast from $m=192$ to the optimum $m=256$. We conclude that a certain number of center points is necessary to cover the point clouds well enough. Once this threshold is reached, the accuracy decreases again.

To study the effect of the varying number of pairwise different center points we also compute the mean, maximum and minimum number of pairwise different center points $m'$ over all objects in an epoch. We see that $m'$ is always significantly smaller than $m$. We conclude that LocAL-Net selects important regions individually for each input point cloud.

\begin{table}[!h]
	\caption{The results of different numbers $m$ of center points on ModelNet40}
	\label{table:center}
	\centering
	\begin{tabular}{l|c|c|c|c|c}
		\#center points $m$ & $192$ & $224$ & $256$  & $288$ & $320$ \\ \hline
		\#mean $m'$ & $125$ & $142$ & $157$  & $173$ & $188$ \\ \hline
		\#max $m'$ & $155$ & $177$ & $202$  & $219$ & $236$ \\ \hline
		\#min $m'$ & $78$ & $86$ & $97$  & $109$ & $122$ \\ \hline
		Acc (\%)   & $92.7$     & $93.2$     & $93.7$ & $93.5$     & $93.4$    
	\end{tabular}
\end{table}

\subsection{Parameter tests}

\textbf{Number of neighbor points.} The local areas are defined by the center points and their neighbors. If the number of neighbors $k$ is too small, a local area does not contain enough information to describe an important region. However, if $k$ is too big the important informations get blurred with unnecessary points. The test results, shown in Table \ref{table:neighbor}, confirm this assumption. LocAL-Net extracts information from local areas most effectively for $k=128$ on ModelNet40.

\begin{table}[!h]
	\caption{The results of different numbers $k$ of neighbor points on ModelNet40}
	\label{table:neighbor}
	\centering
	\begin{tabular}{l|c|c|c|c|c}
		\#neighbor points $k$ & $64$ & $96$ & $128$  & $160$ & $192$ \\ \hline
		Acc (\%)              & $92.8$  & $93.1$  & $93.7$ & $93.1$  & $93.0$
	\end{tabular}
\end{table}

\textbf{Metric features.} In this test we discuss the effects of the metric features on the test accuracy. We set up test scenarios for all combinations of the three features. The results are summarized in Table \ref{table:mf}. It is interesting to note that the individual metric characteristics do not have a large influence on the test result. Only the combination of all three characteristics provides a significant improvement.

\begin{table}[!h]
	\caption{The results of different combinations of the metric features on ModelNet40}
	\label{table:mf}
	\centering
	\begin{tabular}{l|c|c|c|c}
		& $\phi_{j_1}$ & $\phi_{j_2}$ & $\phi_{j_3}$ & Acc (\%) \\ \hline
		A &            &              &              & $93.1$        \\ 
		B & $\checkmark$      &             &             & $93.2$         \\ 
		C &             & $\checkmark$      &             & $93.1$         \\ 
		D &             &             & $\checkmark$      & $93.1$         \\ 
		E & $\checkmark$      & $\checkmark$      &         & $93.2$       \\ 
		F & $\checkmark$      &             & $\checkmark$      & $93.2$        \\ 
		G &             & $\checkmark$      & $\checkmark$      & $93.3$        \\ 
		H & $\checkmark$      & $\checkmark$      & $\checkmark$      & $93.7$  
	\end{tabular}
\end{table}

\begin{table*}[h]
	\caption{Results of the shape part segmentation on ShapeNet.}
	\label{table:seg}
	\centering
	\resizebox{\textwidth}{!}{
		\begin{tabular}{l|c|c|cccccccccccccccc}
			method     & input   & \begin{tabular}[c]{@{}c@{}}instance\\ mIoU\end{tabular} & aero          & bag           & cap           & car           & chair         & \begin{tabular}[c]{@{}c@{}}ear\\ phone\end{tabular} & guitar        & knife         & lamp          & laptop        & \begin{tabular}[c]{@{}c@{}}motor \\ bike\end{tabular} & mug           & pistol        & rocket        & \begin{tabular}[c]{@{}c@{}}skate \\ board\end{tabular} & table         \\ \hline
			Kd-Net \cite{kdtree}     & 4k      & 82.3                                                    & 80.1          & 74.6          & 74.3          & 70.3          & 88.6          & 73.5                                                & 90.2          & 87.2          & 81.0          & 94.9          & 57.4                                                  & 86.7          & 78.1          & 51.8          & 69.9                                                   & 80.3          \\
			PointNet \cite{pn}  & 2k      & 83.7                                                    & 83.4          & 78.7          & 82.5          & 74.9          & 89.6          & 73.0                                                & 91.5          & 85.9          & 80.8          & 95.3          & 65.2                                                  & 93.0          & 81.2          & 57.9          & 72.8                                                   & 80.6          \\
			3DmFV \cite{fisher}  & 2k  & 84.3   & 82.0          & 84.3          & 86.0          & 76.9          & 89.9          & 73.9       & 90.8          & 85.7          & 82.6          & 95.2          & 66.0         &94.0          & 82.6          & 51.5          & 73.5         & 81.8   \\
			KCNet  \cite{kcnet}    & 2k      & 84.7                                                    & 82.8          & 81.5          & 86.4          & 77.6          & 90.3          & 76.8                                                & 91.0          & 87.2          & 84.5          & 95.5          & 69.2                                                  & 94.4          & 81.6          & 60.1          & 75.2                                                   & 81.3          \\
			DGCNN  \cite{dgcnn}   & 2k      & 85.2                                                    & 84.0          & 83.4          & 86.7          & 77.8          & 90.6          & 74.7                                                & 91.2          & 87.5          & 82.8          & 95.7          & 66.3                                                  & 94.9          & 81.1          & 63.5          & 74.5                                                   & 82.6          \\
			Point2Seq \cite{p2s}  & 2k      & 85.2                                                    & 82.6          & 81.8          & 87.5          & 77.3          & 90.8          & 77.1                                                & 91.1          & 86.9          & 83.9          & 95.7          & 70.8                                                  & 94.6          & 79.3          & 58.1          & 75.2                                                   & 82.8          \\
			PointCNN \cite{DBLP:journals/corr/abs-1801-07791}  & 2k      & 86.1                                                    & \textbf{84.1} & 86.5 & 86.0          & \textbf{80.8} & 90.6          & 79.7                                                & \textbf{92.3} & 88.4 & 85.3          & \textbf{96.1} & \textbf{77.2}                                         & \textbf{95.3} & \textbf{84.2} & \textbf{64.2} & \textbf{80.0}                                          & 83.0          \\
			RS-CNN   \cite{rscnn}  & 2k      & \textbf{86.2}                                           & 83.5          & 84.8          & \textbf{88.8} & 79.6          & \textbf{91.2} & 81.1         & 91.6          & 88.4 & \textbf{86.0} & 96.0          & 73.7                                                  & 94.1          & 83.4          & 60.5          & 77.7                                                   & 83.6 \\
			Ours       & 2k      & \textbf{86.2}   & \textbf{84.1}             & \textbf{88.8}             & 86.7             & 78.8             & \textbf{91.2}             & \textbf{83.3}     & 91.9             & \textbf{88.6}          & 84.9          & 95.7      & 72.5      & 94.8             & 83.6             & 60.0             & 77.1        & \textbf{84.0}             \\ \hline
			SO-Net  \cite{sonet}   & 1k, nor & 84.6                                                    & 81.9          & 83.5          & 84.8          & 78.1          & 90.8          & 72.2                                                & 90.1          & 83.6          & 82.3          & 95.2          & 69.3                                                  & 94.2          & 80.0          & 51.6          & 72.1                                                   & 82.6          \\
			PointNet++ \cite{pn++} & 2k, nor & 85.1                                                    & 82.4          & 79.0          & 87.7          & 77.3          & 90.8          & 71.8                                                & 91.0          & 85.9          & 83.7          & 95.3          & 71.6                                                  & 94.1          & 81.3          & 58.7          & 76.4                                                   & 82.6          \\
			SpiderCNN \cite{spidercnn} & 2k, nor & 85.3                                                    & 83.5          & 81.0          & 87.2          & 77.5          & 90.7          & 76.8                                                & 91.1          & 87.3          & 83.3          & 95.8          & 70.2                                                  & 93.5          & 82.7          & 59.7          & 75.8                                                   & 82.8         
		\end{tabular}
	}
\end{table*}

\textbf{Voting.} Finally, we use a voting strategy by averaging the output scores of 10 randomly anisotropic scaled point clouds in the range $[0.66,1.4]$ for each object as proposed in \cite{rscnn}. This test shows the potential of LocAL-Net, which thereby achieves a test accuracy of $93.7\%$. But even without the voting strategy LocAL-Net achieves an accuracy of $93.3\%$ which is the highest result compared to methods equally using only 1,024 points with $xyz$ coordinates (see Table \ref{table:modelnet}).

\subsection{3D Shape part segmentation}

For the 3D Shape part segmentation we evaluate LocAL-Net on the ShapeNet dataset \cite{Yi16}. The dataset contains 16,881 shapes from 16 categories, labeled in a total of 50 parts. The objects are labeled in 2-5 parts. Similar to \cite{pn, pn++} we split the dataset in 14,034 samples for training and 2,847 samples for testing and sample 2,048 points on the mesh of each CAD model. As seen in equation (\ref{seg}) the part segmentation task is formulated as a per-point classification task. The augmentation is the same as for the shape classification task, with the change that the anisotropic scaling is in the range $[0.5, 2.0]$. Similar to PointNet++ we propagate features from center points to the original points by interpolating. We use inverse distance weighted average based on the $k$-NN. 

Let $p_j\in P \setminus P'$ and $\{p'_1,\dots,p'_i,\dots,p'_k\}\subset P'$ the $k$-NN of $p_j$ in $P'$. With $w_i(p_j) = (\|p'_i-p_j\|_2^2)^{-1}$ the feature vector $f$ of $p_j$ is calculated by
\begin{equation*}
	f(p_j) = \dfrac{\sum_{i=1}^{k}w_i(p_j)\tilde{f}(p'_i)}{\sum_{i=1}^{k}w_i(p_j)},
\end{equation*}
where $\tilde{f}(p'_i)$ is the corresponding high-level feature vector of the center point $p'_i$. 
For the evaluation we use Intersection-over-Union (IoU) as metric. For each part of an object, the number of correctly classified points is divided by the union of the predicted points with the target points to obtain the part IoU. To obtain the IoU for an input point cloud the average of all part IoUs is calculated. Finally, the mean IoU (mIoU) is calculated by averaging the IoUs of all input point clouds. Similar to \cite{rscnn} we do a voting with 10 anisotropic scaled point clouds in the test.

We increase the size of the high-level features to $256$ and decreas the size of the global feature $g_2$ to $512$. The metric features, calculated in MFC did not improve the result of the part segmentation test accuracy. Therefore we omit the MFC part to increase the performance. Furthermore, we use one interpolation layer to propagate the high-level feature vectors of the center points $p'_i\in P'$ to the original points $p_j\in P \setminus P'$. Since the interpolation is based on $k$-NN it make sense to use equally distributed center points. Therefor, we use FPS for the segmentation task and increase the number of center points to $m=512$. We still compute $g_1$. Similar to PointNet and PointNet++ we concatenate the one-hot vector indicating the class of the input point cloud and additionally the global feature vector $g$ and the point-features resulting from the CPL part to the output of the last feature layer. 

The results of our segmentation are shown in Table \ref{table:seg}. LocAL-Net is competitive for the task of part segmentation. Moreover, it could reproduce or even improve the best test accuracy in 6 out of 16 categories. The main advantages of LocAL-Net for the segmentation task are the point-features and $g_1$ from the CPL part. 

\section{Conclusion}

In this paper we have proposed LocAL-Net, an architecture for which focuses on learning critical points with a meaningful local neighborhood. Our experiments show that LocAL-Net rarely picks points in flat areas but learns points on the contour and extremity points. Even more, the number of selected neighborhoods depends on the underlying geometry. With this idea LocAL-Net is able to focus on important local areas. We enrich each local area by metric properties which improve learning. In the critical point extraction process the whole point cloud is considered, which additionally provides global features that we use for the classification and segmentation part. Experiments show that these ideas help LocAL-Net to learn locERROR: System command execution is disabled (see Preferences)l and global patterns. Due to this combination LocAL-Net is competitive for the task of part segmentation. For classification LocAL-Net outperforms the state-of-the-arts.

\section{Acknowledgments}

This work was supported by the "Research at Universities of Applied Sciences" programme of the German Federal Ministry of Education and Research, funding code 03FH010IX6.

\bibliographystyle{plain}
\bibliography{LocALNet}

\begin{thebibliography}{10}

\bibitem{pointwise}
Sai-Kit~Yeung Binh-Son~Hua, Minh-Khoi~Tran.
\newblock Pointwise convolutional neural networks.
\newblock {\em Conference on Computer Vision and Pattern Recognition}, pages
  974--993, 2018.

\bibitem{pn++}
Hao~Su Charles Ruizhongtai~Qi, Li~Yi and Leonidas~J. Guibas.
\newblock {PointNet++}: Deep hierarchical feature learning on point sets in a
  metric space.
\newblock {\em Advances in neural information processing systems}, pages
  5099--5108, 2017.

\bibitem{pn}
Kaichun~Mo Charles Ruizhongtai~Qi, Hao~Su and Leonidas~J. Guibas.
\newblock {PointNet}: Deep learning on point sets for 3d classification and
  segmentation.
\newblock {\em Conference on Computer Vision and Pattern Recognition}, pages
  77--85, 2016.

\bibitem{voxandmul}
Matthias Nie{\ss}ner-Angela Dai Mengyuan~Yan Charles Ruizhongtai~Qi, Hao~Su and
  Leonidas~J. Guibas.
\newblock Volumetric and multi-view cnns for object classification on 3d data.
\newblock {\em Conference on Computer Vision and Pattern Recognition}, pages
  5648--5656, 2016.

\bibitem{3dcapsule}
Ali Cheraghian and Lars Petersson.
\newblock 3dcapsule: Extending the capsule architecture to classify 3d point
  clouds.
\newblock {\em Winter Conference on Applications of Computer Vision}, pages
  1194--1202, 2019.

\bibitem{3dcnn}
Rob Fergus-Lorenzo~Torresani Du~Tran, Lubomir D.~Bourdev and Manohar Paluri.
\newblock {C3D:} generic features for video analysis.
\newblock {\em arXiv:1412.0767}, 2014.

\bibitem{octnet}
Ali Osman~Ulusoy Gernot~Riegler and Andreas Geiger.
\newblock Octnet: Learning deep 3d representations at high resolutions.
\newblock {\em Conference on Computer Vision and Pattern Recognition}, pages
  863--872, 2016.

\bibitem{mvcnn}
Evangelos~Kalogerakis Hang Su~and, Subhransu~Maji and Erik~G. Learned{-}Miller.
\newblock Multi-view convolutional neural networks for 3d shape recognition.
\newblock {\em International Conference on Computer Vision}, pages 945--953,
  2015.

\bibitem{pweb}
Chi-Wing~Fu Hengshuang~Zhao, Li~Jiang and Jiaya Jia.
\newblock Pointweb: Enhancing local neighborhood features for point cloud
  processing.
\newblock {\em Conference on Computer Vision and Pattern Recognition}, pages
  5560--5568, 2019.

\bibitem{learnvox}
Tianfan Xue William T.~Freeman Jiajun~Wu, Chengkai~Zhang and Joshua~B.
  Tenenbaum.
\newblock Learning a probabilistic latent space of object shapes via 3d
  generative-adversarial modeling.
\newblock {\em Advances in neural information processing systems}, pages
  82--90, 2016.

\bibitem{rotnet}
Asako Kanezaki.
\newblock Rotationnet: Learning object classification using unsupervised
  viewpoint estimation.
\newblock {\em Conference on Computer Vision and Pattern Recognition}, pages
  5010--5019, 2018.

\bibitem{kdtree}
Roman Klokov and Victor~S. Lempitsky.
\newblock Escape from cells: Deep kd-networks for the recognition of 3d point
  cloud models.
\newblock {\em International Conference on Computer Vision}, pages 863--872,
  2017.

\bibitem{sonet}
Gim Hee~Lee Lee~Jiaxin, LiBen M.~Chen.
\newblock So-net: Self-organizing network for point cloud analysis.
\newblock {\em Conference on Computer Vision and Pattern Recognition}, page
  9397–9406, 2018.

\bibitem{adam}
Kingma~Diederik P. and Ba~Jimmy.
\newblock Adam: A method for stochastic optimization.
\newblock {\em International Conference on Learning Representations}, pages
  1--13, 2015.

\bibitem{p2s}
Yu{-}Shen Liu Matthias~Zwicker Xinhai~Liu, Zhizhong~Han.
\newblock Point2sequence: Learning the shape representation of 3d point clouds
  with an attention-based sequence to sequence network.
\newblock {\em Association for the Advancement of Artificial Intelligence},
  2019.

\bibitem{voxfpnn}
Hao Su Charles Ruizhongtai~Qi Yangyan~Li, S{\"o}ren~Pirk and Leonidas~J.
  Guibas.
\newblock {FPNN:} field probing neural networks for 3d data.
\newblock {\em Advances in neural information processing systems}, page
  307–315, 2016.

\bibitem{DBLP:journals/corr/abs-1801-07791}
Mingchao~Sun Yangyan~Li, Rui~Bu and Baoquan Chen.
\newblock {PointCNN}: Convolution on x-transformed points.
\newblock {\em Advances in neural information processing systems}, page
  828–838, 2018.

\bibitem{Yi16}
Li~Yi, Vladimir~G. Kim, Duygu Ceylan, I-Chao Shen, Mengyan Yan, Hao Su, Cewu
  Lu, Qixing Huang, Alla Sheffer, and Leonidas Guibas.
\newblock A scalable active framework for region annotation in 3d shape
  collections.
\newblock {\em SIGGRAPH Asia}, 2016.

\bibitem{spidercnn}
Mingye Xu Long Zeng Yu~Qiao Yifan~Xu, Tianqi~Fan.
\newblock Spidercnn: Deep learning on point sets with parameterized
  convolutional filters.
\newblock {\em European Conference on Computer Vision}, page 90–105, 2018.

\bibitem{kcnet}
Yaoqing Yang Dong~Tian Yiru~Shen, Chen~Feng.
\newblock Mining point cloud local structures by kernel correlation and graph
  pooling.
\newblock {\em Conference on Computer Vision and Pattern Recognition}, page
  4548–4557, 2018.

\bibitem{fisher}
Michael~Lindenbaum Yizhak~Ben{-}Shabat and Anath Fischer.
\newblock 3d point cloud classification and segmentation using 3d modified
  fisher vector representation for convolutional neural networks.
\newblock {\em arXiv:1711.08241}, 2017.

\bibitem{rscnn}
Shiming~Xiang Yongcheng~Liu, Bin~Fan and Chunhong Pan.
\newblock Relation-shape convolutional neural network for point cloud analysis.
\newblock {\em Conference on Computer Vision and Pattern Recognition}, pages
  8895--8904, 2019.

\bibitem{dgcnn}
Ziwei Liu Sanjay E. Sarma Michael M.~Bronstein Yue~Wang, Yongbin~Sun and
  Justin~M. Solomon.
\newblock Dynamic graph {CNN} for learning on point clouds.
\newblock {\em arXiv:1801.07829}, 2018.

\bibitem{mn}
Aditya Khosla Fisher Yu Linguang Zhang Xiaoou~Tang Zhirong~Wu, Shuran~Song and
  Jianxiong Xiao.
\newblock {3D ShapeNets}: A deep representation for volumetric shapes.
\newblock {\em Conference on Computer Vision and Pattern Recognition}, pages
  1912--1920, 2015.

\end{thebibliography}

\end{document}